\documentclass[letterpaper, 10 pt, conference]{ieeeconf}  

\IEEEoverridecommandlockouts                              

\usepackage[table]{xcolor}  

\usepackage{graphicx} 
\usepackage{todonotes}
\usepackage{algorithm}
\usepackage[noend]{algpseudocode}
\usepackage{amsmath}  
\usepackage{amssymb}
\usepackage{longtable}
\usepackage{url}
\usepackage[abs]{overpic}

\usepackage{listings}
\usepackage{xspace}

\usepackage{xspace}

\newcommand{\ie}{i.\,e.,\xspace}

\newcommand{\sv}[1]
{{\normalfont\textsf{\small #1}}} 
\newcommand{\smallsv}[1]{{\fontsize{7pt}{7pt}\textsf{#1}}} 
\newcommand{\RAE}{\sv{RAE}\xspace}
\newcommand{\UPOM}{\sv{UPOM}\xspace}
\newcommand{\RU}{\sv{RAE+UPOM}\xspace}

\newcommand{\phead}[1]{
	\\ \multicolumn{3}{@{}l}{#1}}

\newcounter{pline}

\newcommand{\1}{\\&}

\newcommand{\pkey}[1]{\\#1:&}

\newenvironment{pcode}{%
	\setcounter{pline}{0}%
	\begin{tabular}[b]{@{\kern .8em}r@{~}l@{}l@{\,}}
		&&\\[-5.5ex] 
	}{%
	\end{tabular}%
}

\newlength{\tabsize}
\newcommand{\T}{\hspace*{\tabsize}}

\title{\LARGE \bf Acting and Planning with Hierarchical Operational Models\\ on a Mobile Robot: A Study with RAE+UPOM}

\author{
Oscar Lima$^{1,2,*}$,
Marc Vinci$^{1,2,*}$,
Sunandita Patra$^{3,*}$,
Sebastian Stock$^{1}$,\\
Joachim Hertzberg$^{2,1}$,
Martin Atzmueller$^{2,1}$,
Malik Ghallab$^{4}$,
Dana Nau$^{3}$,
Paolo Traverso$^{5}$
\thanks{This work is supported by the German Federal Ministry of Education and Research (BMBF). The DFKI Niedersachsen Lab (DFKI NI) is sponsored by the Ministry of Science and Culture of Lower Saxony and the VolkswagenStiftung.}%
\thanks{$^{1}$ German Research Center for Artificial Intelligence (DFKI),}
\thanks{Cooperative and Autonomous Systems, Osnabrück, Germany,}
\thanks{\{oscar.lima, marc.vinci, sebastian.stock\}@dfki.de}%
\thanks{$^{2}$ Osnabrück University, Institute of Computer Science, Germany}
\thanks{\{joachim.hertzberg, martin.atzmueller\}@uos.de}%
\thanks{$^{3}$ University of Maryland, College Park, USA, \{patras, nau\}@umd.edu}%
\thanks{$^{4}$ LAAS-CNRS, Toulouse, France, malik@laas.fr}%
\thanks{$^{5}$ Fondazione Bruno Kessler (FBK-ICT), Italy, traverso@fbk.eu}%
\thanks{* These authors contributed equally to this work.}%
}

\begin{document}
\maketitle
\thispagestyle{empty}
\pagestyle{empty}

\definecolor{custompink}{RGB}{255, 20, 147}
\begin{abstract}
Robotic task execution faces challenges due to the inconsistency between symbolic planner models and the rich control structures actually running on the robot. In this paper, we present the first physical deployment of an integrated actor-planner system that shares hierarchical operational models for both acting and planning, interleaving the Reactive Acting Engine (\smallsv{RAE}) with an anytime UCT-like Monte Carlo planner (\smallsv{UPOM}). We implement \smallsv{RAE+UPOM} on a mobile manipulator in a real-world deployment for an object collection task.
Our experiments demonstrate robust task execution under action failures and sensor noise, and provide empirical insights into the interleaved acting-and-planning decision making process.

\vspace{0.2cm}
{\color{custompink}\normalfont\url{https://dfki-ni.github.io/RAE_ECMR_2025}}
\end{abstract}

\section{Introduction}
\label{sec:intro}
In robotics tasks, synthesizing an effective plan of actions traditionally relies on descriptive models, which abstractly specify nominal outcomes of actions to compute state transitions efficiently. However, executing these planned actions requires operational models, where rich control structures and closed-loop decision-making adapt to real-time, nondeterministic execution outcomes. Operational models are essential for handling events and situation dynamics. 

Typically, when integrating task planning and plan execution methods on a mobile robot, there is a gap between the descriptive model (\ie the symbolic model with preconditions and effects, such as PDDL) used by the planner and the operational  model (\ie in concrete implementations) that is physically executed on the robot~\cite{ingrand2017deliberation}. The plans generated by the task planner are executed on the robot by a plan execution module that iterates over the plan and dispatches actions based on the results of previous actions; and it triggers re-planning in case of execution failures. Descriptive models use simplifying assumptions in order to speed up the planning process.
However, concrete implementations of actions on the robot can contain complex control structures in software code and algorithms, e.g., motion planning, as well as dedicated low-level error handling.

This gap between the planner’s descriptive models and the robot’s operational models introduces challenges in aligning symbolic assumptions with the behavior of low-level control structures, which are further exacerbated by dynamic events.

The \RU~\cite{patra2021deliberative} system was built to address these challenges, where both acting and planning share a set of hierarchical operational models with rich control structures.
\RAE (Refinement Acting Engine) selects actions by using an integrated UCT probabilistic Planner for Operational Models (\UPOM) and interleaves it with action execution, following the principles outlined in~\cite{ghallab2016automated}.
While \RAE was designed for general-purpose applications and demonstrated on modeled robotic use cases, these were limited to toy domains and have not been deployed on real robotic systems. 

\definecolor{darkolive}{rgb}{0.2, 0.3, 0.1}
\definecolor{light-gray}{gray}{0.75}
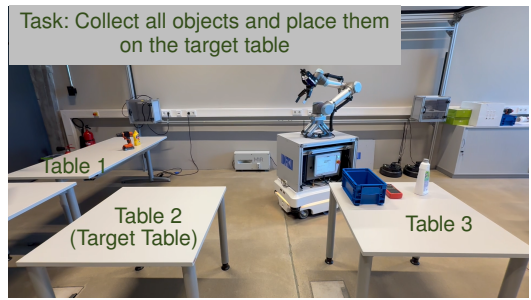
\begin{figure}[t]
    \centering
    \begin{overpic}[width=0.8\linewidth]{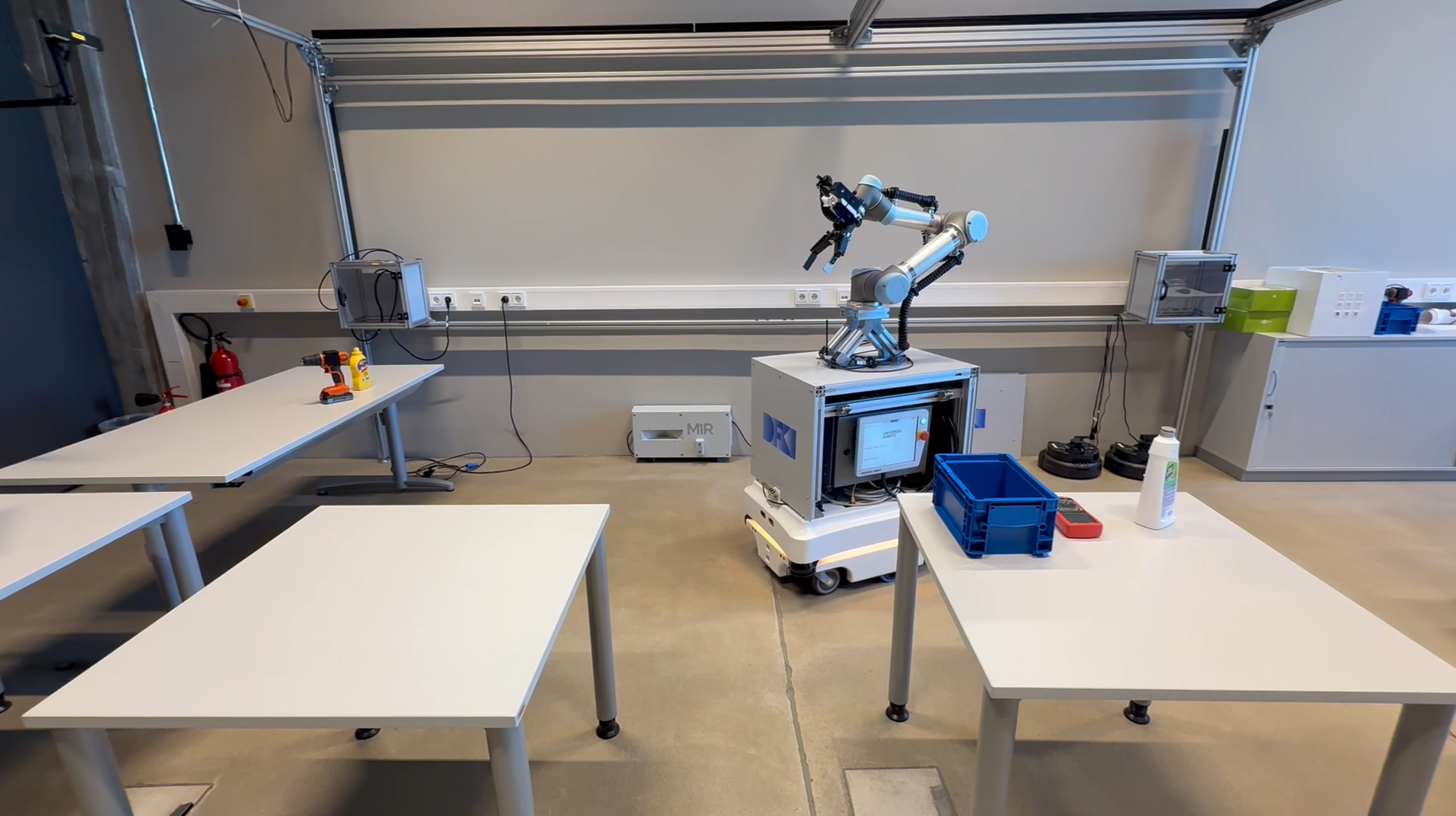}
        \put(0,97){\colorbox{light-gray}{\parbox{5cm}{\centering \sffamily \footnotesize \textcolor{darkolive}{Task: Collect all objects and place them on the target table}}}}
        \put(23,20){\sffamily\footnotesize\textcolor{darkolive}{(Target Table)}}
        \put(12,47){\sffamily\footnotesize\textcolor{darkolive}{Table 1}}
        \put(40,28){\sffamily\footnotesize\textcolor{darkolive}{Table 2}}
        \put(150,25){\sffamily\footnotesize\textcolor{darkolive}{Table 3}}
    \end{overpic}
    \caption{Mobipick robot setup at the beginning of most real-world trials, without prior knowledge of object poses. Table identifiers correspond to those used in our experimental trials.}
    \label{fig:real_mobipick_env}
    \vspace{-0.5cm}
\end{figure}

This work presents the first real-world deployment of \RU on a physical robot - the mobile manipulator Mobipick  (Fig.~\ref{fig:real_mobipick_env}). We evaluate the integrated system's performance in a practical setting and analyze some challenges that arose during execution. This empirical assessment provides insights into the strengths and limitations of the approach, contributing to the ongoing discourse on integrated acting and planning in robotics.
The Mobipick robot consists of a mobile base, a custom middle-part and a UR5 arm with a camera (see~\cite{lima2023physics} for a detailed description of the robotic platform). We model pick-and-place tasks for moving objects between tables overcoming navigation and perception failures. The robot operates in a fully autonomous manner, leveraging state-of-the-art open-source software. 

Our main contributions are: (a) adaptation and extension of the \RU engine for integration with a physical robotic platform, enabling robust operation under real-world execution constraints, (b) hierarchical operational model for object collection, shared between actor and planner that enables the robot to perceive and collect objects effectively, and, (c) experiments on a real robot in a controlled hardware lab environment, providing insights into the planner's internal decision-making mechanisms.

\section{Related Work}
\label{sec:sota}
The popular 3-layered architecture combines, with different representations, (i) a low level for the sensori-motor control, (ii) an execution level, and (iii) a planning level~\cite{ingrand2017deliberation}. Our approach integrates levels (ii) and (iii) using hierarchical operational models, while still needing level (i) control routines, that are paired on the robot and the simulator.

Several systems aim to integrate planning and execution within robotic frameworks. ROSPlan~\cite{cashmore2015} connects task planning and dispatching with a custom symbolic knowledge base in ROS, distributing plan generation and execution across multiple nodes. PlanSys2~\cite{martin2021} partially replicates this approach in ROS2, introducing Behavior Trees for plan execution as a key enhancement. The Embedded Systems Bridge (ESB)~\cite{sadanandam2023closed} is a middleware agnostic Python library for robotic systems that simplifies the creation of domain models for the Unified Planning (UP)~\cite{micheli2025unified} library and connects representations for sensor readings and execution code with their planning domain counterpart. It executes and monitors plans based on this mapping. 
However, these systems keep planning and execution knowledge separate.
In contrast, we use operational models for both planning and acting. To our knowledge, we are the first to use hierarchical operational models for integrated acting and planning on a real robot.

There are planners combining hierarchical controllers with planning (e.g., OMPAS~\cite{turi2022extending}, Behavior Trees~\cite{colledanchise2019towards}, TAMP planners~\cite{garrett2021integrated}). Like our approach, they target the problem of closed-loop online decision making by leveraging these control structures.
Turi et al.~\cite{turi2023enhancing} propose an acting system with time and resource management (OMPAS) and integrates an online hierarchical planner to continuously guide the actor. OMPAS also builds upon \RAE and is an acting system that uses hierarchical models to perform multiple high-level tasks simultaneously by executing commands on a robotic platform. OMPAS manages resources required by skills during execution. In contrast to our approach it uses a dedicated acting language, SOMPAS, whereas in our approach, the domain is directly written in Python code. The system demonstrates improved robustness in simulations (i.e., factory tasks), but not on physical robots. \RU performs simulated execution of the code to reason about the outcomes of choosing one method for a task over another.

\section{Modeling Object Collection Using Hierarchical Operational Models}
\label{sec:modeling}
In this section, we define the object collection task and describe how we model it using refinement methods.

\newcommand{\current}{i}
The robot operates in a hardware lab environment (Fig. \ref{fig:real_mobipick_env}), where object instances from a known set of types are distributed across multiple tables. The robot's task is to identify and transport objects to a designated target table. Each object type appears only once in this environment, eliminating the need for persistent object anchoring across perception steps. The robot has no prior knowledge of the objects' locations on the tables and must handle action failures, sensor noise, and state uncertainty, all influenced by the objects' physical and visual properties. The robot's manipulation is limited by its workspace, specifically reachability constraints. Based on perceptual input, our system can generate symbolic facts about whether an object is \textit{on} a table or \textit{in} a box. \cite{lima2023physics}. Each table holds various objects characterized by utility values and success probabilities for collection, which together influence the prioritization of object retrieval. Among these is a box object, which carries no utility itself but can be used to transport multiple items simultaneously. Guided by deliberative acting with \RU, the robot perceives its environment and determines the collection order to maximize the expected utility. To achieve this, it must actively explore the tables, devise collection strategies, and adapt to events, including navigation and perception failures.

\subsubsection*{\textbf{Collection Task}}
Given a set of objects,
    $O = \{ o_i \}_{i=1}^{N}$,
scattered across locations in unknown poses, and a robot with perception, manipulation, and navigation capabilities, the goal is to collect a subset $O' \subseteq O$ and place them at one of the designated target tables,
$
    T = \{ tb_i \}_{i=1}^{p},
$
in a limited amount of time (a discrete counter), optimizing a utility function based on object rewards and action costs (Eq~\ref{eq:utility}). We define action costs using discrete time steps. We use the same definition of the collection task in \RU, the simulated collection scenario, and the real-robot demonstrator. A hierarchical operational model is a set of {\em refinement methods}, which are computer programs giving alternative ways of performing tasks. A refinement method has the form:

\begin{center}
\small\begin{pcode}
\phead{method-name($\textit{arg}_1$, $\textit{arg}_2$, ..., $\textit{arg}_k$)}
\pkey{task} task or event identifier, $t$($\textit{arg}_1$, $\textit{arg}_2$, ..., $\textit{arg}_l)$
\pkey{pre} test for preconditions
\pkey{body} program
\pkey{params} expression to get all possible values of \textit{\{arg$_j$\}$_{j=l+1}^k$} 
\end{pcode}
\end{center}

The method for a task $t$ specifies {\em how} to perform $t$, i.e., it gives a procedure for accomplishing $t$ by performing sub-tasks, commands, and state variable assignments. The procedure may include any of the usual programming constructs: if-else statements, loops, etc. A method can have more arguments than the task. The {\em params} expression defines the range of such uninstantiated arguments. A {\em method instance} is a method with instantiated parameters. Method instances differ in their parameter values and give \RU alternative ways to do the same task. An example refinement method for collecting objects from a table is shown below:

\vspace{0.3cm}
{\rm\small
\begin{pcode}
\phead{\sv{collect\_objs\_from\_table}$(r : \text{robot}, tb : \text{table})$}
\pkey{task}{\sv{collect\_objs\_from\_table}$(r: \text{robot})$}
    \pkey{body}if $P_B(r)$ $\neq$ base($tb$): \sv{drive}($r$, base($tb$))
    \1 while \sv{time\_passed} $\leq \textit{LIMIT}$:
    \1 \T objects\_on\_table $\gets$ \{$o$ $|$ on($o$)=$tb$, $C_O$($o$)$\neq$box\}
    \1 \T if objects\_on\_table $\neq \emptyset$:
    \1 \T \T \sv{collect\_obj}($r$, $tb$)
    \1 \T \T \sv{drive}($r$, base($tb$))
    \1 \T else: break
    \1 $b \gets$ $\{b |$\smallsv{on}($b$, $tb$) $\wedge$ $C_O(b) = $ box $\wedge$ \smallsv{in}$(obj', b)\}$
    \1 if $b$ $\neq$ None : \sv{move\_object}($r$, $b$, $tb$, \textit{target\_tb})
\pkey{params} $\{tb \in T | \text{\sv{not\_visited}($tb$)} \wedge tb \neq \textit{target\_tb}$\}
\end{pcode}
}%

\begin{figure}[t]
\centering
\includegraphics[width=\columnwidth]{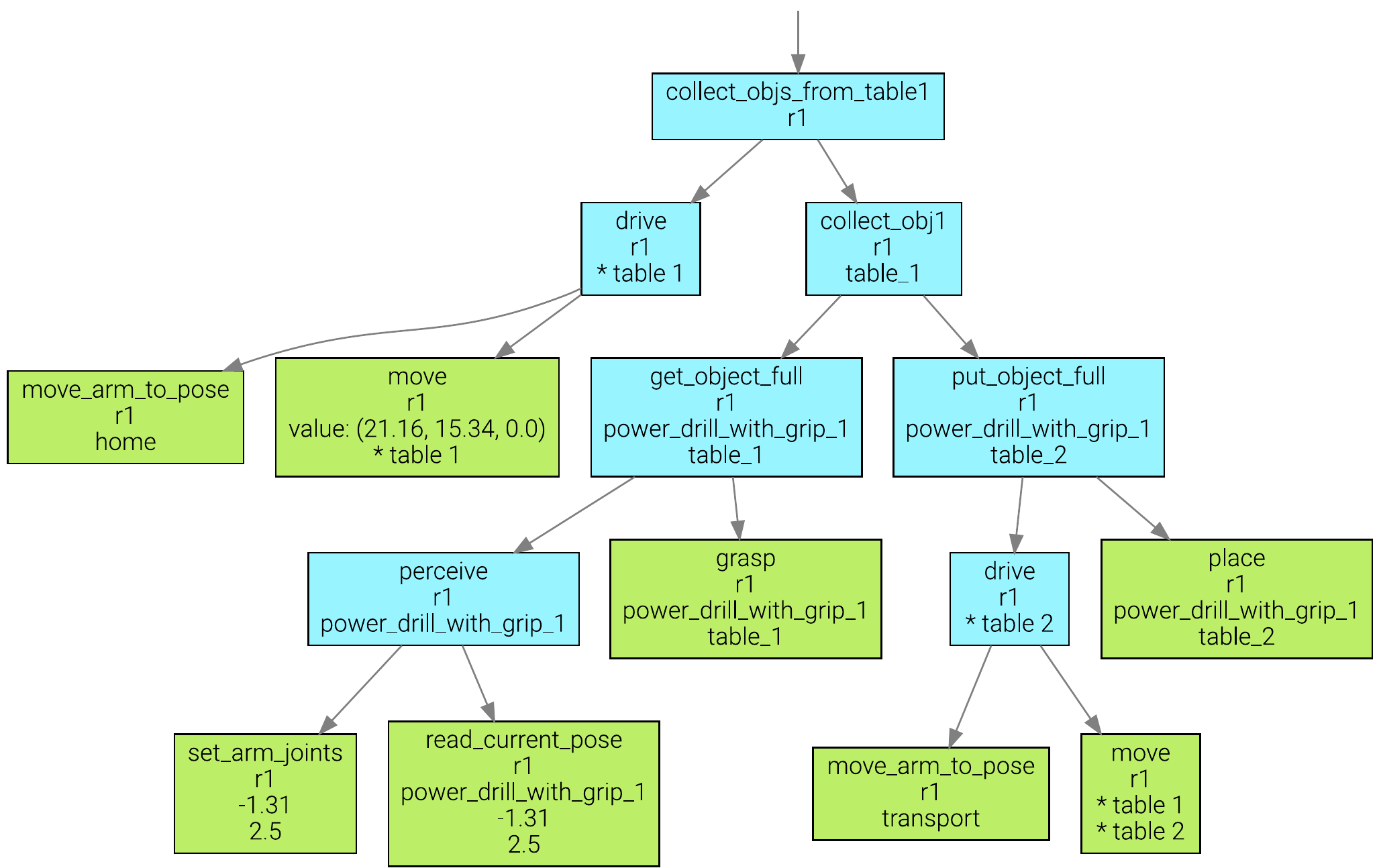}
\caption{A dynamically \smallsv{UPOM}-generated refinement tree for a \textit{specific} collection task \smallsv{collect\_objs\_from\_table}$(r1)$ for the current state.
}
\label{fig:refinement-tree}
\vspace{-0.4cm}
\end{figure}

In Fig.~\ref{fig:refinement-tree} we present one of the possible refinement trees resulting from the execution of \sv{collect\_objs\_from\_table}$(r1, tb1)$, where \sv{drive}$(r1, tb1)$ and \sv{collect\_obj}$(r1, tb1)$ are dynamically generated subtasks. We have two methods for the \sv{collect\_obj} task, one that involves the use of a box, and another that does not.
Formally, we define a domain as a tuple $\Sigma=(S, \mathcal{T, M, A})$ where $S$ is the set of states, $\mathcal{T}$ is the set of tasks that the robot needs to accomplish, $\mathcal{M}$ is the set of method instances for handling tasks in $\mathcal{T}$, and $\mathcal{A}$ is the set of commands that can be executed by the robot. Due to limited space, we summarize the set of state variables, tasks, and commands in Table~\ref{table:domain}. We then describe the refinement methods used for object perception.

{\small{
\begin{table}[htbp]
\centering
\centering
\begin{tabular}{|l|p{6.1cm}|}
\hline
\textbf{State variables} & \textbf{Description} \\
\hline
$P_B$ & Pose of the robot, defined as $P_B = (x, y, \theta)$, where $(x, y)$ are 2D coordinates and $\theta$ is the orientation. \\
\hline
$P_O$ & Object poses, where $P_O(o) = (x, y, \theta)$ for each object $o \in O$. Initialized as \textit{unknown} before perception. \\
\hline
$P_{sym}$ &  \smallsv{on}$(obj, table)$, \smallsv{in}$(obj, box)$: Symbolic facts derived from object and robot poses. 
\\
\hline
$C_O$ & Classification of each object into predefined categories (e.g., tool, box, table). \\
\hline
 \smallsv{holding} & State of the robot's manipulator—whether it is holding an object and which one. Verified manually as it may be incorrect (e.g., if the object is dropped). \\
\hline
\smallsv{not\_visited}($tb$) & whether table $tb$ is visited or not\\
\hline
\smallsv{collected}($tb$) & set of collected objects in table $tb$\\
\hline
\end{tabular}
\smallskip

\begin{tabular}{|p{0.97\linewidth}|}
\hline
\textbf{Tasks:}
{\smallsv{explore}(\textit{r})}, {\smallsv{collect\_obj}(\textit{r}, \textit{tb})}, {\smallsv{perceive}(\textit{r}, \textit{o})}, {\smallsv{drive}(\textit{r}, \textit{l})}, {\smallsv{collect\_objs\_from\_table}(\textit{r})}, {\smallsv{get\_object}(\textit{r},~\textit{o},~\textit{tb})}, {\smallsv{put\_object}(\textit{r}, \textit{o}, \textit{tb})}, {\smallsv{move\_object}(\textit{r}, \textit{o}, \textit{from}, \textit{to})}, {\smallsv{insert\_object}(\textit{r}, \textit{tb},~\textit{o},~\textit{box})}, {\smallsv{collect\_all\_objs}(\textit{r})} \\
\hline
\end{tabular}
\smallskip

\begin{tabular}{|p{0.97\linewidth}|}
\hline
\textbf{Commands (low level actions):}
{\smallsv{move}(\textit{r}, \textit{l1}, \textit{l2})}, {\smallsv{read\_current\_pose}(\textit{r}, \textit{o}, \textit{pan}, \textit{lift})}, {\smallsv{grasp}(\textit{r}, \textit{o}, \textit{from\_what})}, {\smallsv{perceive\_table}(\textit{r}, \textit{tb}, \textit{pan}, \textit{lift})}, {\smallsv{move\_arm\_to\_pose}(\textit{r}, \textit{arm\_pose})}, {\smallsv{set\_arm\_joints}(\textit{r}, \textit{pan}, \textit{lift})}, {\smallsv{place}(\textit{r}, \textit{o}, \textit{on\_what})}, {\smallsv{store\_object}(\textit{r}, \textit{tb}, \textit{o}, \textit{box})} \\
\hline
\end{tabular}
\caption{A summary of the state variables, tasks and commands. } 

\label{table:domain}
\vspace{-0.75cm}
\end{table}}}

\subsection{Modeling the Perception Task}

Hierarchical refinement methods with nondeterministic actions are well-suited for modeling and reasoning about perception tasks. We assume no prior knowledge of object poses and define methods to estimate the pose of an object by exploring different viewpoints using its camera. 

Perception is modeled in two steps: first, a coarse estimate of all objects is obtained using an \sv{explore} task; later, when an object becomes a collection target, a fine-grained perception step is performed to accurately estimate its pose and ensure successful grasping.

\subsubsection*{\textbf{Exploring items on all tables}}

Initially, an \sv{explore}($r$) task iterates over the tables, navigating to each and performing a wide-angle scan from a distance using a predefined arm posture and camera configuration. This coarse estimation populates the state with approximate object poses, enabling \RU to prioritize tables with more objects to maximize utility. The \sv{perceive\_table} command is then used to detect all visible objects from that configuration.

\subsubsection*{\textbf{Perception before grasping}} Once the robot moves closer to the target table, fine perception becomes necessary, as localization noise introduces pose uncertainty. Therefore, before grasping, the robot refines the object pose estimate by selecting a suitable camera pose (adjusting pan and lift joints) to ensure complete object visibility. \RU can guide the perception by selecting values for two specific arm joints, pan and lift, while the remaining joints are fixed to position the camera toward the center of the tables. The refinement method for the \sv{perceive}(\textit{r, o}) task is defined as follows:

\vspace{0.1cm}
\begin{center}
		{\rm\small
			\begin{pcode}
	    	\phead{\sv{perceive\_method}$(r, o, \textit{pan, lift})$}
				\pkey{task}{\sv{perceive}$(r, o)$}
				    \pkey{body}\sv{set\_arm\_joints}($r$, \textit{pan, lift})
				    \1 if dist($P_B$($r$), $P_O$($o$)) $\leq$ $d_{MAX}$:
				    \1 \T     \sv{read\_current\_pose}($r$, $o$, \textit{pan, lift})
        \pkey{params}\textit{pan} $\in$ range(\textit{SHOULDER PAN JOINT})
        \1 \textit{lift} $\in$ range(\textit{ELBOW LIFT JOINT})
			\end{pcode}
		}
\end{center}

The parameters \textit{pan} and \textit{lift} take values from discrete sets defined by a human expert, representing the range of motion for the respective joints that fully cover the largest table.

In the method, the robot adjusts its joints using the \sv{set\_arm\_joints} command. If the distance between the robot's pose and the object's pose (estimate from initial \sv{explore} task described earlier) is within the maximum specified range for pose estimation $d_{MAX}$, the robot retrieves the object's pose via \sv{read\_current\_pose}.

\begin{figure}
    \centering
    \begin{minipage}{0.49\linewidth}
        \centering
        \includegraphics[width=\linewidth]{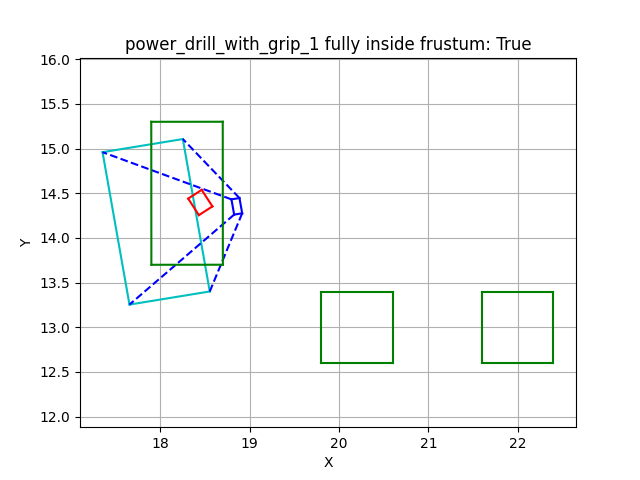}
    \end{minipage}
    \hfill
    \begin{minipage}{0.49\linewidth}
        \centering
        \includegraphics[width=\linewidth]{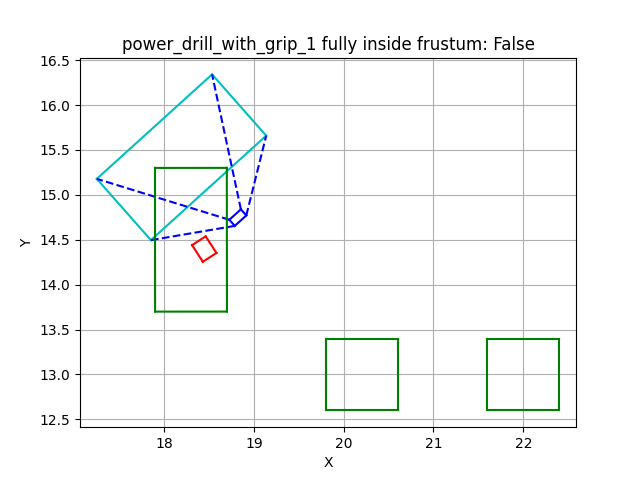}
    \end{minipage}
    \caption{Visualization of the object visibility check function used for the simulation of \smallsv{read\_current\_pose} command. Left: the object detection is likely to be successful as it is fully inside the camera frustum. Right: detection fails due to an inadequate camera pose that leaves the object out of sight.}
    \label{fig:fov_example}
    \vspace{-0.5cm}
\end{figure}

During planning, the \sv{read\_current\_pose} command verifies whether a specified object is fully contained within the robot’s camera frustum. Because the camera is mounted on the robot's end-effector, its pose can be derived from joint encoder values. After the initial exploration provides approximate object poses and known bounding boxes, the function computes the camera's frustum volume in 3D space and verifies whether all bounding box vertices reside within this volume. The command succeeds only if the object is entirely contained within the frustum; otherwise, it fails. As illustrated in Fig.~\ref{fig:fov_example}, this visibility check enables reliable object pose estimation by guiding the selection of camera poses that maximize the likelihood of observation during planning. At execution time, we call DOPE~\cite{tremblay2018corl:dope}, a deep-learning model that processes camera images and outputs the object's 6D pose estimate and class.

For planning, we use a simplified 2D object pose representation (x, y, $\theta$), where $\theta$ denotes the top-down orientation of the object, and all other pose dimensions are discarded.

\section{Refinement Acting and Planning}
\label{sec:ref_act_plan}
\begin{figure}[b]
    \centering
    \includegraphics[width=\linewidth]{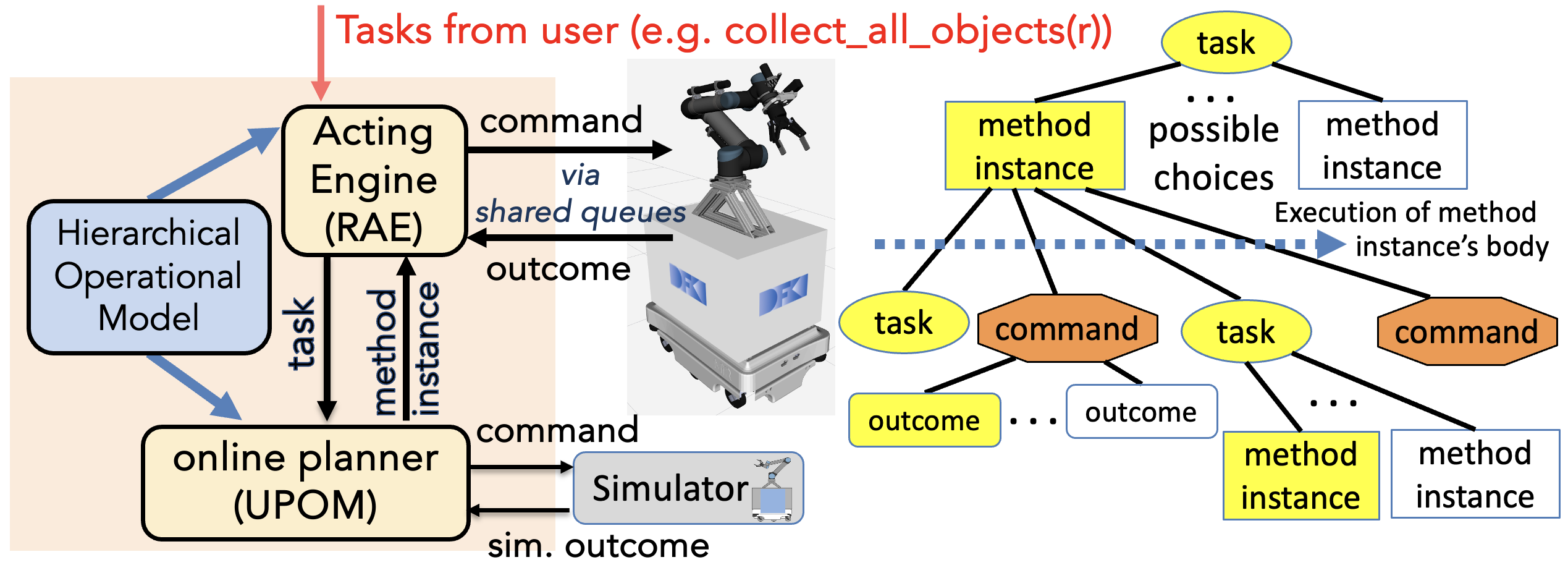}
    \caption{Left: schematic diagram of our acting and planning architecture integrated with a robot. Right: possible choices of method instances for a task. Each command has an associated utility and \smallsv{UPOM} suggests a suitable method instance based on expected utility.}
    \label{fig:schematic}
\end{figure}

This section provides an overview of \RU, which comprises two main subsystems: \RAE, the acting engine responsible for executing refinement methods, and \UPOM, the planner that estimates the effectiveness of each method in context. For further information about \RU and convergence guarantees, refer to~\cite{patra2021deliberative}. Fig.~\ref{fig:schematic} illustrates the architecture as implemented in our system.

A task may have several refinement method instances (method + parameters), each describing a different way to perform it. The best method instance to use depends on the specific state. \UPOM performs Monte Carlo rollouts, which are simulated executions of applicable method instances in the current state. The rollouts use probabilistic action outcomes given by a simulated hand-crafted model. \UPOM performs a recursive search to find a method instance $m$, for a task $\tau$ in a state $s$, approximately optimal for a utility function. The accuracy of its recommendations improves with the number of rollouts. This reduces discrepancies between planning and acting that stem from differences in method evaluation and execution dynamics. 

\subsection{Interleaving Acting and Planning}
We state the deliberative acting problem as: given $\Sigma=(S, \mathcal{T, M, A})$ and a task (e.g., \sv{collect\_obj\_from\_table}($r1$)) $\tau \in \mathcal{T}$, find the ``best'' method instance $m \in \mathcal{M}$ to accomplish $\tau$ in a current state $s$. The actor requires an online selection procedure which designates for each task or sub-task at hand the best method \textit{with method parameters} to pursue in the current context. When the arguments of a method are not arguments of the task that the method is used for, there are combinatorially many different ways for \RAE to instantiate these parameters when creating method instances. \UPOM does an UCT-like search over the possible method parameters, and suggests the method instance (method + parameter values) with the highest expected utility. Each time a task/sub-task is refined, \UPOM chooses the values of the methods' uninstantiated parameters online.

\RAE uses \UPOM to select a method $m$ and begins executing its steps. For each step $m[i]$, if the command is already running on the robot, \RAE checks its status. If the command is still in progress, \RAE waits or attends to other pending tasks. If the command fails, it calls the \sv{Retry} procedure to explore alternative methods for the current sub-task. Upon successful completion of the command, \RAE proceeds to the next step in the method.

Compared to our previous implementation~\cite{patra2021deliberative}, we made the following two updates to \RU: (1) When a command fails, the utility of a rollout is the sum of the utilities from all successfully collected objects until that point, and not zero. This ensures that utility accounts for partial success when a command fails after collecting some objects. (2) We added a retry count parameter to the methods, allowing multiple retries of the same method before it is discarded. This is useful, e.g., in navigation tasks, where a method that fails once may still succeed upon retry.

\subsubsection*{\textbf{Utility Function}} To promote early collection of high-value objects, the utility function assigns higher estimates to plans that retrieve them sooner. Each object $o \in O$ has an associated reward, $R: O \rightarrow \mathbb{R}^+$, When an object 
$o$ is successfully collected, its base reward $R(o)$ is added to the utility, scaled by an exponentially decaying factor to favor early collection. Let $\langle s_0, a_1, s_1, ..., s_{n-1}, a_n, s_n\rangle$ be the execution trace of a rollout ($s_i \in S, a_i \in \mathcal{A}$), and \textit{cost}$(s, a, s')$ be the cost of performing action $a$ in state $s$ when the outcome
is $s'$.
We define the utility of a rollout as:
\begin{equation}\textstyle
U = \sum_{i=1}^n R(\sv{collected\_at}(s_i)) \cdot \left( c_1 + c_2 \cdot e^{-k \cdot C_i} \right),
\label{eq:utility}
\end{equation}
\noindent where, 
$C_i = \sum_{j=1}^i \textit{cost}(s_{j-1}, a_{j-1}, s_j)$,
 is the cumulative cost up to step $i$,
$\sv{collected\_at}(s_i)$ is the object successfully put on the target table in step $i$. $R(\sv{collected\_at}(s_i))$ is 0 if no object is collected in step $i$. $k$ is the decay constant, and $c_1$, $c_2$ are scaling constants with $c_1 + c_2 = 1$. For each rollout, \UPOM  ensures $C_n \leq \eta$, the time limit. We model time using a discrete counter.

\subsection{Integration with the robotic platform}
Communication between the robot and \RAE is managed through three shared queues. Users submit new tasks via the \textit{task queue}, where each task remains until \RAE retrieves it and selects an applicable refinement method. After planning, \RAE executes the refinement method and places the resulting commands into the \textit{command execution queue} for the robot to execute. Commands are low level actions that can be executed on the robot (see Section ~\ref{sec:modeling}). Upon completion, the robot reports the outcome and its updated state via the \textit{command status queue}, which \RAE monitors to maintain an accurate internal state.

\subsubsection*{\textbf{Executable and Simulated Control Routines}}

Each low-level control routine executed on the robot is paired with a simulated counterpart, enabling \UPOM to perform rollouts using predicted outcomes (Fig.~\ref{fig:schematic}, box `Simulator'). These routines are associated with cost estimates and probabilistic outcomes provided by human experts. While these estimates may be inaccurate, e.g., due to hardware-specific variability, they should at least be sufficiently realistic for relative comparison. The simulation code for each routine updates the system state and returns \emph{success} or \emph{failure}, modeling the expected effects of the routine on the robot. Preconditions are embedded in the operational model (either in a command function or refinement method) and are checked by evaluating the current state, returning \emph{failure} if the routine is not applicable. Control routines are nondeterministic, and the predictive models currently used by \UPOM mainly sample from probability distributions to capture this uncertainty. 
While this integrated actor–planner design removes the model/implementation gap, it does add upfront effort: beyond authoring operational methods we must supply a lightweight simulation layer and a success-detection mechanism for every low-level routine.

\section{Experimental Evaluation}
\label{sec:eval}
We conducted three experimental studies to evaluate our approach’s performance on the object collection task.
Each experiment involves transporting objects from two source tables to a target table (Fig. \ref{fig:real_mobipick_env}). Every object type in the planning model is associated with a utility value and an estimated probability of successful collection, reflecting perception and grasp reliability. These probabilities were empirically determined and combined into a single metric to capture their interdependence, which was found acceptable for our object set. By design, the box is assigned no utility, but it serves as a transport container for other objects.

Each experiment utilized 100 \UPOM rollouts for method and parameter selection per subtask, achieving fast planning times of $\sim 1$~sec. while consistently selecting suitable methods and parameters. One of our objectives was to evaluate whether \RAE can select actions that maximize the expected utility under uncertainty. Fig.~\ref{fig:real_mobipick_env} shows a typical experimental configuration. While further details are provided later, the robot was generally expected to prioritize the table with the highest expected utility, first collecting and delivering high-value objects, and using the box when available. \UPOM evaluated multiple rollouts and, in most cases, selected the object-table pair with the highest expected utility. Table~\ref{table:combined_trial_info_and_results} presents the collected utility, planning times, and overall runtime for each trial.

To systematically evaluate our approach, we organized the experiments into three studies, each focusing on a different aspect of the robot's performance. Each study comprises one or more trials, where a trial denotes a single execution of the object collection task under specific conditions. We first present results from trials demonstrating successful task completion under nominal conditions, followed by trials examining the performance under particular spatial constraints and execution failures.

\begin{table}[t]
  \centering
  \scriptsize
  \begin{tabular}{|l|c|c|c|c|}
    \hline
    \textbf{Trial Description} & \textbf{\shortstack{Collected\\Utility}} & \textbf{\shortstack{Coll.\\Objs}} & \textbf{\shortstack{\rule{0pt}{1.0em}Planning\\Time}} & \textbf{\shortstack{Acting\\Time}} \\
    \hline
    \rowcolor{brown!10}
    \textbf{\rule{0pt}{1.0em}1.1 Box used - high utility} & \textbf{47.45} & 4/4 & 42.18s & 500s \\
    \rowcolor{brown!10}
    \textbf{1.2 Box unavailable} & 32.09 & 4/4 & 40.74s & 549s \\
    \rowcolor{brown!30}
    \textbf{2.1 Spread out objects} & 27.69 & 2/4 & 21.55s & 359s \\
    \rowcolor{brown!20}
    \textbf{3.1 Arm failed} & 27.49 & 3/4 & N/A & 396s \\
    \rowcolor{brown!20}
    \textbf{3.2 Navigation failed} & 35.83 & 3/4 & 28.77s & 453s \\
    \rowcolor{brown!20}
    \textbf{3.3 Perception failed} & 32.73 & 3/4 & 28.59s & 425s \\
    \hline
  \end{tabular}
  \caption{Trial results: utility, collected objects, times.}
  \label{table:combined_trial_info_and_results}
  \vspace{-0.85cm}
\end{table}

\subsubsection*{\textbf{Study 1 – Task completion with high utility}}
In both trials, the robot successfully completed the object collection task, attaining high utility and retrieving all objects. In particular, we focused on the impact of using a transport box on task execution. In trial 1.1, the robot utilized the box to transport multiple objects, whereas in trial 1.2, the box was intentionally unavailable, requiring the robot to transport objects individually. Using the box enabled the robot to complete the task $\sim49$ seconds faster and achieve a higher expected utility of 47.45, compared to 32.1 without the box. While the box itself was assigned zero utility, its behavior is modeled through dedicated refinement methods and symbolic fluents. Its use is not directly incentivized by the utility function (Eq.~\ref{eq:utility}) but emerges through planning when it enables more efficient object transport.

\subsubsection*{\textbf{Study 2 – Increased Object Spacing}} Trial 2.1 evaluated the robot's performance in object collection under conditions of increased spatial separation between objects. The objective was to encourage the robot to adopt alternative, task-appropriate observation poses that ensured all relevant objects were within the field of view. The results aligned with the expectations: the robot successfully selected suitable observation poses to accommodate the dispersed object layout, without requiring any changes to the underlying model or execution platform.

In this trial, the robot failed to perceive the objects on Table 3 and was therefore unaware of their existence. As a result, it could not reason or plan for their collection. This type of failure corresponds to the perception failure scenario described in Trial 3.3.

\subsubsection*{\textbf{Study 3 – Execution failure}} These trials examined failures related to perception, navigation, and arm movement.

Trial 3.1 – Cable Entanglement: The robot has a cable routed from the end-effector to the onboard PC, which can occasionally become entangled during arm movements, triggering the emergency stop mechanism. In this trial, such an entanglement occurred, prompting the operator to quickly intervene and recover the system. During the interruption, \RAE continued retrying the actions until it was able to resume and finish the task.

Trial 3.2 – Navigation Failure: The navigation module failed to reach the initially targeted table. In response, \RAE selected and retried a different table, enabling the trial to complete successfully. As described in Section~\ref{sec:ref_act_plan}, the system retries failed actions using the same parameters, up to a predefined retry count. In this trial, the navigation retry count was set to 2, resulting in 3 failed attempts to reach the target location. This failure was not treated as permanent; the robot later returned to the same table and successfully collected objects. The remaining uncollected object was due to a perception failure on Table 3, unrelated to navigation.

Trial 3.3 – Perception Failure: The detection of the mustard object is highly sensitive to the camera’s viewing angle and distance. If the detection confidence drops below a threshold of 0.5, the object is not recognized at all, which was the case in this trial. Consequently, \RAE ignores the object, as it is unaware of its existence in such cases. At present, such cases remain unhandled and addressing them would require a different modeling approach to improve robustness against undetected objects.

\begin{figure}
    \centering
    \includegraphics[width=\linewidth]{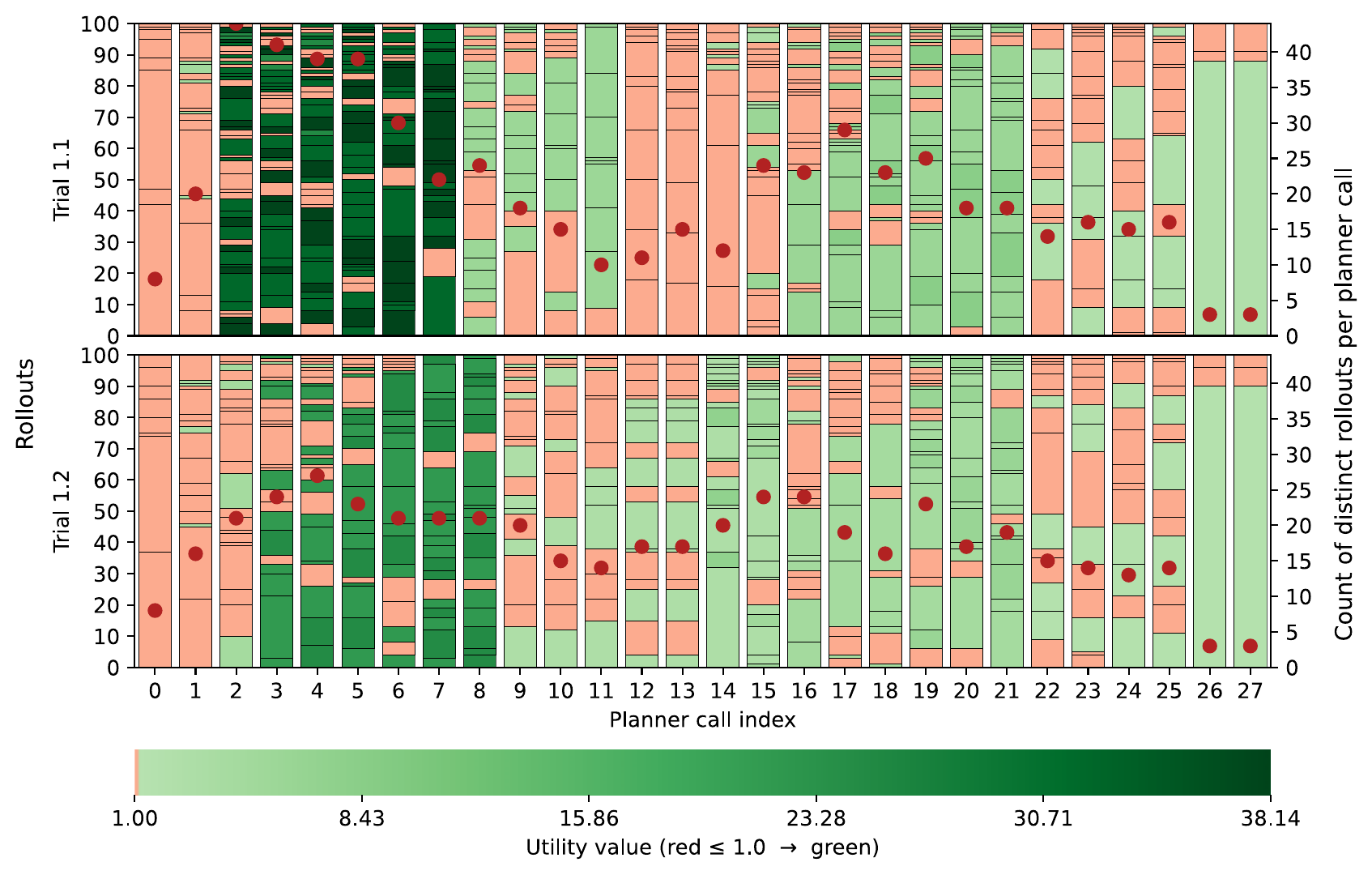}
    \caption{Heatmap of rollout utility across trials 1.1 and 1.2 (with and without box). Color encodes outcome (red = failure, green = success, shade $\propto$ utility), and bar height reflects how often each rollout was simulated. Red dots indicate the number of unique rollout clusters per planner call.}
    \label{fig:combined_heatmap_summary}
    \vspace{-0.5cm}
\end{figure}

Fig.~\ref{fig:combined_heatmap_summary} shows a heatmap-style visualization of planner rollouts from Study 1, generated automatically from planning logs. The plot provides insights into the planner’s behavior, where each row represents a trial and each column corresponds to a planner call during that trial. Color shading indicates the utility of the rollouts: green indicates successful planning outcomes (i.e., utility $>1.0$), while red indicates failure, meaning no utility was found and thus the result contributes no information to \RAE. This highlights a critical aspect of \sv{RAE}’s behavior when planning fails: it defaults to choosing one of the untried refinement methods by \RAE.
Note that high-utility segments correspond to rollouts that successfully reach and evaluate applicable actions, which are then executed by the actor in subsequent steps.

Each stacked bar segment in the plot represents a rollout container, which is a cluster of identical paths in the search space. Because some of \UPOM's rollouts may be identical (exploitation), containers vary in size according to the frequency of identical paths. Larger containers indicate that a particular set of parametrized actions was simulated many times, whereas smaller containers reflect one-off or infrequent evaluations. The number of rollout clusters per planner call, indicated by red dots, provides insight into the diversity of explored strategies. Fewer, larger clusters often indicate a constrained search space or high redundancy, whereas a large number of small clusters suggest extensive exploration and greater variability in planning outcomes.

Across trials, early planner calls consistently yielded no utility, reflecting the necessity of object discovery. Since the utility function is defined to assign no reward to exploration and perception tasks, initial rollouts terminate before reaching utility-contributing steps, resulting in a predominance of red segments. As the planner gathers information and adapts, subsequent calls often show increased success rates, with a visible shift toward green and darker shades—particularly in trials that achieve early high-utility rollouts, benefiting from the system’s time-decaying reward structure.

\section{Conclusions}
\label{sec:conclusions}
Our work demonstrates the usefulness of adapting the \RU acting and planning framework to a real world robot, specifically in the context of a hierarchical object collection task using operational models. With our approach, the robot remains always reactive to its environment while systematically planning for its actions. We empirically demonstrated that we can model perception and manipulation in settings characterized by partial observability, uncertainty, and action failures. Unlike conventional approaches in robotics, which are often rigid, difficult to maintain or reuse, and lack systematic failure handling, our approach enhances reactivity and facilitates maintainability, reusability, and robustness.

In future work, we plan to investigate how the system handles dead ends, more dynamic events and failure types, as well as using learning for refinement method selection within \RAE and prioritizing rollouts via heuristics.

\bibliographystyle{IEEEtran}
\bibliography{references}

\end{document}